\documentclass[10pt]{article}

\usepackage[margin=0.85in]{geometry}
\usepackage{times}
\usepackage{booktabs}
\usepackage{microtype}
\usepackage[hidelinks]{hyperref}

\title{\vspace{-2.2em}\textbf{When Is a Multi-Agent Code Judge Actually Grounded?}\\
\large Two Label-Free Measurements, and a Judge That Declines to Guess\vspace{-0.6em}}
\author{\normalsize
Salma Roshdy Aly \quad Hussein Assaf \quad Ziad Kobti \\[0.25em]
\small School of Computer Science, University of Windsor, Canada \\[0.15em]
\small \texttt{\{ali9o, assafh, kobti\}@uwindsor.ca}}
\date{\vspace{-1.4em}}

\begin{document}
\maketitle
\vspace{-1.2em}

Software teams increasingly let one language model write code and a second decide
whether that code is correct enough to merge \cite{MT-Bench, LLM-as-a-Judge-for-swe}. The second model acts as a safety
check, and it has a failure mode that accuracy alone does not reveal. When it lacks any basis for a decision, it does not report that fact. It returns a confident verdict with reasoning attached, and that verdict is indistinguishable from one it had good
reason for \cite{Relying-on-the-Unreliable}, so the reviewer downstream sees a decision rather than the absence of
evidence behind it.

As discussed in \cite{Can-LLMs-Replace-Human-Evaluators}, one promising response is to stop asking a single model for a verdict. Instead
the judgment is broken into small checkable claims, and a separate agent
verifies each claim against evidence while being kept away from the first model's
reasoning so that it cannot simply agree with it. This works well when the
evidence is a set of retrieved documents. Our question is what the method needs
in order to work at all, and whether that can be checked before anyone relies on
it.

We argue that the evidence must satisfy two conditions. It has to be independent
of the answer under review, since otherwise the checker rereads that answer and
agrees with it, and it has to differ between the two candidates being compared,
since evidence identical for both is perfectly reliable and yet tells
nothing. The first condition is familiar; the second is rarely stated, because
with retrieved documents it holds automatically; code judging is where it stops
holding.

To test this we ran a published verification framework, MARCH \cite{march}, unmodified on a
code judging benchmark \cite{codejudgebench}: two solutions to the same problem, one correct and one
subtly buggy, and the judge must pick the correct one. The framework runs three agents over each solution. A solver reads
the problem and the solution and forms an opinion about whether it is correct. A
proposer turns that opinion into a list of individually checkable claims. A
checker then answers each claim, seeing the problem and the solution but not the
solver's reasoning, so that it cannot restate what it is meant to be verifying.
Whichever solution survives more of its own claims wins. We evaluate on
CodeJudgeBench, whose paired solutions come from three generators
(\textsc{gemini-2.5-pro}, \textsc{claude-3.7-sonnet} and \textsc{qwen3-235b})
across three difficulty levels, and on HumanEvalFix \cite{octopack}, whose defects are written by
hand into short library functions rather than sampled from a model. Judging is
carried out by qwen3 family \cite{qwen3}, specifically, \textsc{qwen3-8b} and repeated with \textsc{qwen3-14b}, giving 54
experimental cells and roughly 2{,}000 comparisons. The table reports
CodeJudgeBench, the benchmark that carries difficulty labels.

The pipeline never fails. Claims are produced, parsed and verified, and
no stage reports an error. Yet it declares the two solutions equally good on 78
to 95\% of comparisons, reaching 4.4\% accuracy where the same model asked
directly reaches 43.7\% ($p = 1.1 \times 10^{-20}$). Easier problems do not
improve it: the direct judge gains up to 49 accuracy points moving from hard
problems to easy ones, while the pipeline gains less than one. Nor does a larger
judge, with 1.75 times the parameters, change the picture. Two measurements
account for this, and both can be computed from logs the pipeline already writes
without any ground truth labels. On 76.6\% of comparisons the system asked the
same questions about both solutions, so the answers could not tell them apart,
and the checker agreed with 80.7\% of the claims it was asked to verify.

The same measurement suggests a repair. If the proposer asked identical questions
about both solutions, the pipeline has no way to separate them, and that can be
seen before the checker is run at all. Withholding those comparisons and letting
the pipeline decide only the rest raises its accuracy from 20.7 to 36.9\% while
still answering half of all comparisons, and the gain holds in every generator
and difficulty level we tested and under both judges. It does not make the
pipeline competitive with the single model it is built from, which answers every
comparison and is right on 73.4\% of them but improves the pipeline.

\vspace{0.1em}
\begin{center}
\footnotesize
\setlength{\tabcolsep}{4pt}
\renewcommand{\arraystretch}{1.0}
\begin{tabular}{lccccc}
\toprule
\textbf{Difficulty} & \textbf{Direct accuracy} & \textbf{MARCH ungated accuracy} & \textbf{Comparisons kept after gating} & \textbf{MARCH gated accuracy} & \textbf{Gain} \\
\midrule
Easy   & 94.3\% & 28.7\% & 62.5\% & 43.5\% & $+14.9$ \\
Medium & 75.0\% & 21.2\% & 54.3\% & 35.4\% & $+14.2$ \\
Hard   & 45.1\% & 13.5\% & 37.9\% & 28.9\% & $+15.4$ \\
All (Aggregated)   & 73.4\% & 20.7\% & 50.6\% & \textbf{36.9\%} & $+16.2$ \\
\bottomrule
\end{tabular}
\end{center}
\vspace{-0.5em}
{\footnotesize
\emph{Direct accuracy} is the same model asked for a verdict without the
pipeline. \emph{Ungated accuracy} is the pipeline's figure over every comparison,
\emph{comparisons kept} the share the gate lets through, the rest being withheld
rather than answered, and \emph{gated accuracy} its figure over those kept. The
gate reads nothing but the proposer's questions, so it needs no labels and no
extra model calls, and it fires before the checker runs.\par}
\vspace{0.2em}

Halving the comparisons attempted and nearly doubling accuracy on the rest is a
partial repair, not a solution. Our contribution
is not a more accurate judge, but a way to tell when a multi-agent judge such as MARCH has no basis for its
answer, drawn from records the pipeline already keeps, and a step that declines
to answer in those cases instead of guessing.

% Our contribution is not a more accurate judge, but a label-free criterion for detecting when a multi-agent judge such as MARCH has no basis for its answer, and a selective-prediction step that declines to answer in those cases instead of guessing. Where standard selective prediction thresholds a confidence estimate~\cite{kamath2020selective}, the gate reads only the pipeline's own records, which is what the setting permits when the score is uninformative by construction.

\bibliography{refs}
\bibliographystyle{abbrv}

\end{document}